\documentclass[a4paper,conference]{IEEEtran}
\ifCLASSINFOpdf
  \usepackage[pdftex]{graphicx}
\else
\fi
\usepackage{amsmath}
\usepackage{bm}
\usepackage{booktabs}
\usepackage{multirow}
\usepackage{hyperref}
\usepackage{lineno}
\usepackage{subfigure}
\usepackage{xcolor}

\begin{document}
%
\title{Universal Model for Multi-Domain Medical Image Retrieval}

\author{\IEEEauthorblockN{
Yang Feng, Yubao Liu and
Jiebo Luo}
\IEEEauthorblockA{Department of Computer Science, University of Rochester, Rochester, NY 14627\\
\{yfeng23, jluo\}@cs.rochester.edu, liuyubao96@gmail.com}}


%


\maketitle

\begin{abstract}
Medical Image Retrieval (MIR) helps doctors quickly find similar patients' data, which can considerably aid the diagnosis process. MIR is becoming increasingly helpful due to the wide use of digital imaging modalities and the growth of the medical image repositories. However, the popularity of various digital imaging modalities in hospitals also poses several challenges to MIR. Usually, one image retrieval model is only trained to handle images from one modality or one source. When there are needs to retrieve medical images from several sources or domains, multiple retrieval models need to be maintained, which is cost ineffective. In this paper, we study an important but unexplored task: how to train one MIR model that is applicable to medical images from multiple domains? Simply fusing the training data from multiple domains cannot solve this problem because some domains become over-fit sooner when trained together using existing methods. Therefore, we propose to distill the knowledge in multiple specialist MIR models into a single multi-domain MIR model via universal embedding to solve this problem. Using skin disease, x-ray, and retina image datasets, we validate that our proposed universal model can effectively accomplish multi-domain MIR.
\end{abstract}


%
\IEEEpeerreviewmaketitle

\section{Introduction}
Healthcare systems are rapidly adopting digitization, making the type and number of medical images stored and accessible more than ever. With a large medical image repository, many applications can be enabled. For example, it is possible to find similar cases to the current patient according to the medical images and then diagnose the current patient by taking similar cases into consideration. Efficiently finding similar cases in large medical image repositories, known as MIR (medical image retrieval), is crucial to the success of digital healthcare.

Existing MIR methods mainly focus on retrieving medical images from one source or one domain. When multiple sources of images are available, the common choice is to train an equal number of MIR models with each model in charge of one source of images. Needless to say, maintaining and serving multiple MIR models would incur high costs and efforts. 

\begin{figure}
\centering
\includegraphics[width=\columnwidth]{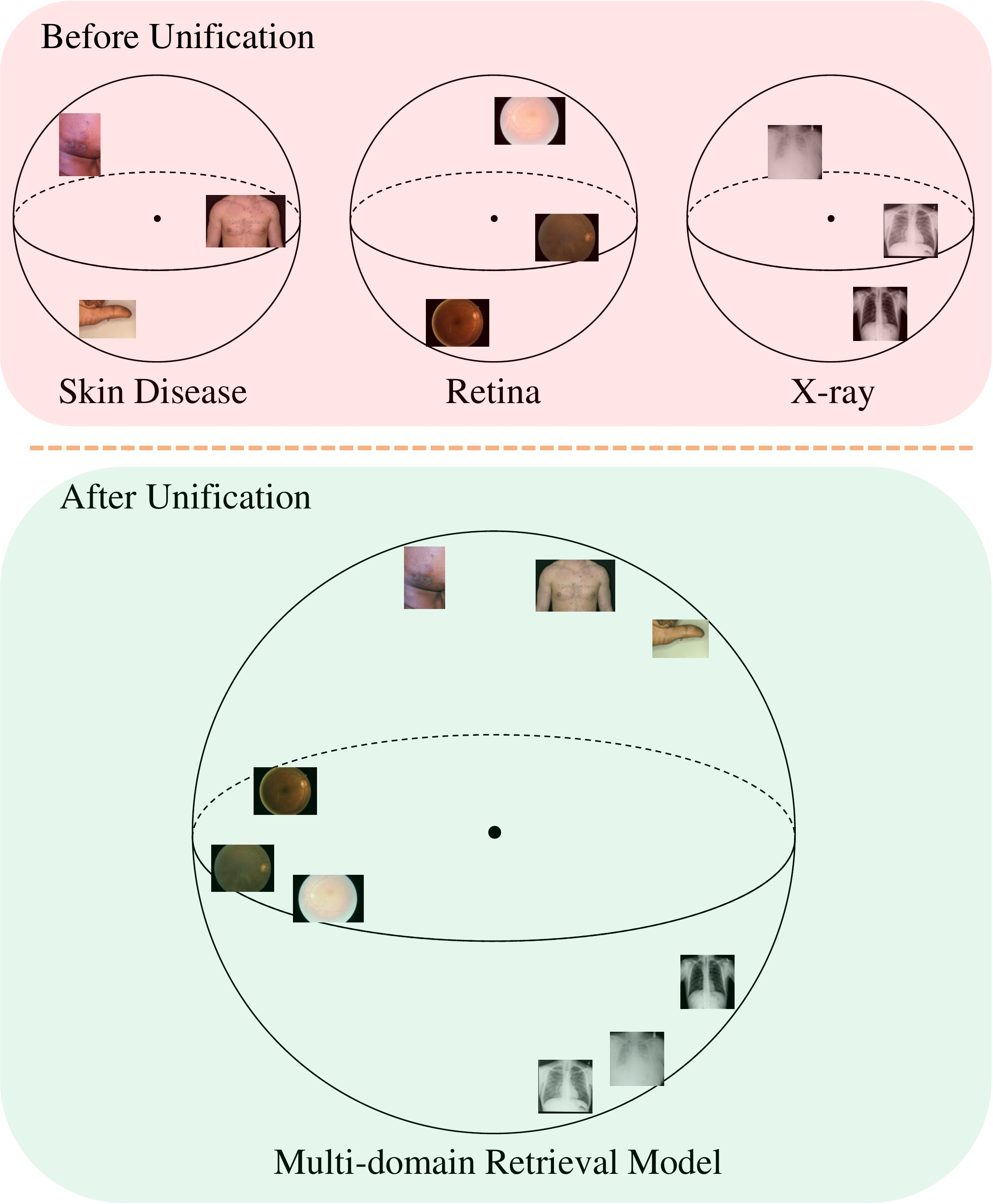}
\caption{Before unification, one MIR model is needed for one source or domain of medical images. We aim to train a single universal MIR model that is applicable to multiple sources of medical images.}
\label{fig:aim}
\end{figure}

In this paper, we aim to train a single universal MIR model that is applicable to multiple sources of medical images. Fig.~\ref{fig:aim} illustrates the idea of multi-source MIR. Multi-source MIR has two advantages. First, only one model needs to be served and maintained, which costs less. Second, more training images will be available after fusing multiple sources, which may potentially improve the retrieval performance of all sources. Compared with natural image retrieval, the available training images for one source of medical image is often much fewer. Therefore, MIR performance would suffer from the lack of training data. It is shown in \cite{zhou2019models} that natural images can also be helpful to medical image applications. Based on their observation, we investigate whether the medical images from one source can help the retrieval of medical images from another source. In the experiments, we have showed that this is exactly the case.

\begin{figure*}[t]
    \centering
\subfigure{
  \includegraphics[width=0.31\textwidth]{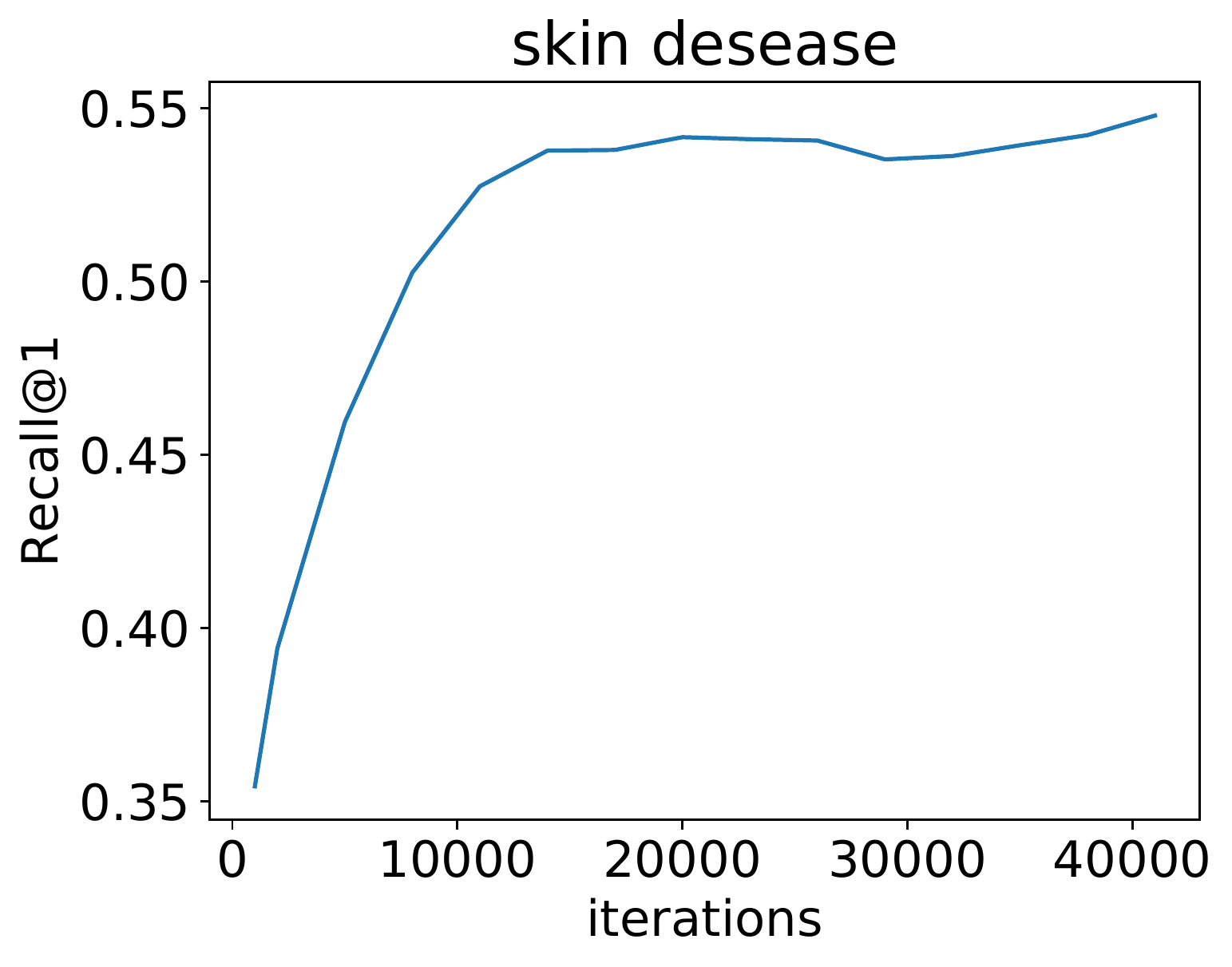}
}
\subfigure{
  \includegraphics[width=0.31\textwidth]{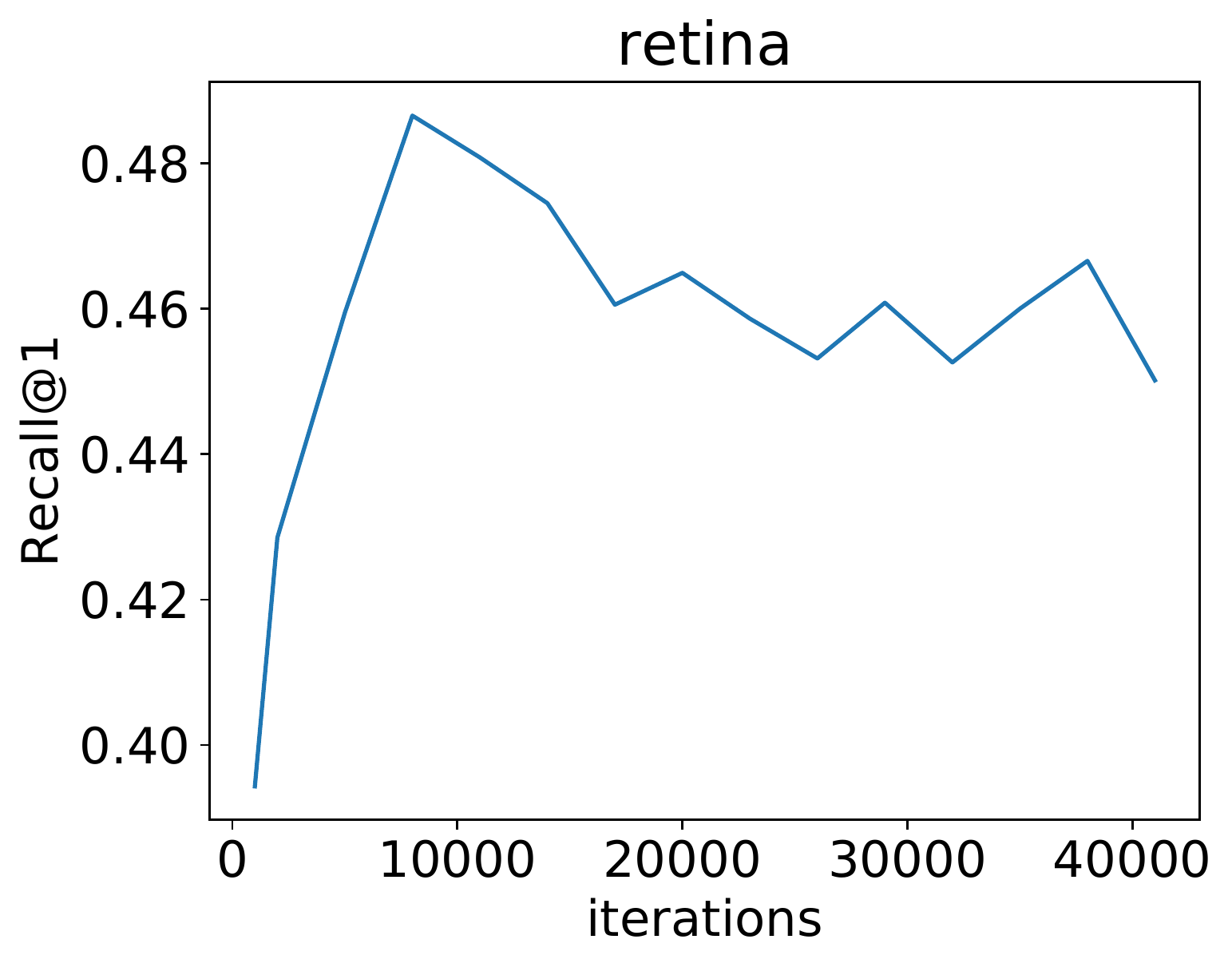}
}
\subfigure{
  \includegraphics[width=0.31\textwidth]{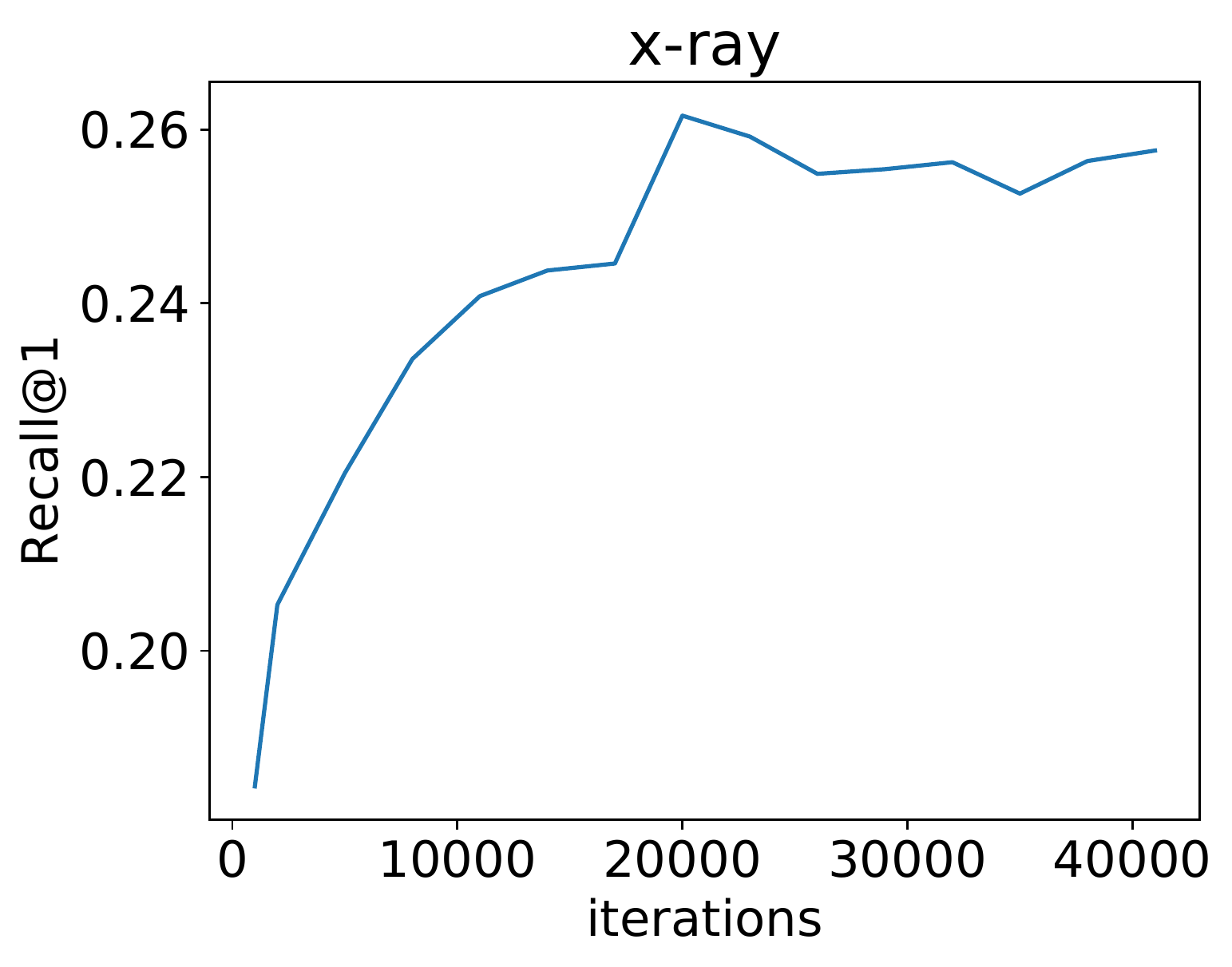}
}
    \caption{Visualization of the Recall@1 values when the skin disease, retina, and x-ray images are fused and trained together. Please refer to Sec.~\ref{sec:exp} for more experiment details. Clearly, the best stopping checkpoints are different for different sources.}
    \label{fig:recall}
\end{figure*}

The na\"ive way to train a multi-source MIR model is to fuse the training images from multiple sources together and then train with existing methods. However, the performance of the resulted model will be worse than a specialist MIR model trained on a given single source of images. We analyze the reason for this phenomenon and find that some sources of images can become over-fit much faster than other sources and thus it is impossible to choose a stopping point where the single MIR model provides the best performance for all the sources, as illustrated in Fig.~\ref{fig:recall}. A second reason is that it becomes more difficult to sample effective training pairs or triplets after fusing images from multiple sources. The images across different sources generally have larger differences than within the same source. For the triplet loss~\cite{weinberger2006distance}, it will be too easy for a training triplet to produce no gradients if the negative image and anchor image are from different sources. These ineffective triplets will form the majority of all the triplets when fusing multiple sources, thus making it more difficult to identify effective triplets during training. One may suspect that data imbalance may also be a reason for the performance decrease in Fig.~\ref{fig:recall}, but we do not believe it is the case because of two facts: 1) the proposed method in this paper also suffers from data imbalance but still achieves good performance; and 2) a data-balanced baseline we add in the experiment also suffers from the early overfitting issue.

To train a universal multi-source MIR model that can match the performance of multiple specialist MIR models on each specialist’s domain, we propose to distill the knowledge learned in all the specialists into a universal multi-source MIR model. By distilling the knowledge from properly trained specialists, the obtained multi-source MIR model will not overfit on any source. The distillation for multi-source MIR is very different from the existing distillation methods~\cite{hinton2015distilling,park2019relational}, which distill the knowledge either from a large teacher model to a smaller student model or from an ensemble of models to a single model. Note that the ensemble consists of models trained on the same data with different initialization. In contrast, the specialist MIR models are trained on different data in this paper. To achieve multi-source MIR, another possible way is a straightforward direct ensemble of all the specialist MIR models. However, the direct ensemble method would require higher computation and achieve lower performance, as will be shown in the experiment.

In summary, our contributions are three-folds: 1) we identify an important but unexplored task: how to train a multi-source MIR model to match the performance of several specialists on each specialist’s domain; 2) we propose to use distillation to avoid the early overfitting of some sources when training the multi-source model; and 3) we validate the effectiveness of the proposed multi-source MIR method in experiments by retrieve the images from skin disease, retina, and x-ray sources.

\section{Related Work}
\textbf{Deep Metric Learning} Recent deep metric learning research mainly falls into three directions: loss function, sampling policy, and learning with proxy. 
Contrastive loss~\cite{chopra2005learning} aims to pull similar sample pairs closer to each other and push dissimilar sample pairs farther away. However, directly minimizing the absolute distances of similar pairs to zero may be too restrictive. Triplet loss~\cite{weinberger2006distance} was proposed to solve this issue. In triplet loss, the relative distance order between anchor-positive and anchor-negative is ensured. Recently, researchers have found that embedding models trained with the Softmax classification loss also have good performances~\cite{qian2019softtriple,sohn2016improved,wen2016discriminative}. A fact about the contrastive loss and triplet loss is that the number of pairs or triplets is quadratic or cubic to the number of training samples, respectively. With so many triplets, it is would be challenging to find an effective triplet to produce gradients if uniformly sampling is used. 
Schroff \textit{et al.}~\cite{schroff2015facenet} proposed semi-hard sampling to find the first negative farther than the positive to the anchor. Wu \textit{et al.}~\cite{wu2017sampling} proposed to sample negative samples reciprocal to the anchor-negative distance. They showed that their sampling policy leads to a more balanced distribution of anchor-negative distance. Duan \textit{et al.}~\cite{duan2019deep} designed a deep sampler network to learn the sampling strategy. Instead of sampling positives and negatives from a large pool of candidates, Movshovitz \textit{et al.}~\cite{movshovitz2017no} proposed to use a proxy to represent a class of samples. Both the training speed and model performance were improved by introducing proxies.

Existing research on deep metric learning mainly focuses on training a specialist embedding model for one source of images. We are interested in training a multi-source retrieval model having good performance on multiple sources.

\textbf{Knowledge Distillation} Initially, Bucilua \textit{et al.}~\cite{bucilua2006model} compressed large ensemble models into smaller and faster models. The ensemble was used to label a large unlabeled dataset. Thereafter, the small neural network was trained using the ensemble labeled data. Hinton \textit{et al.}~\cite{hinton2015distilling} improved the compression method by introducing a temperature to reduce the effect of large negative logits. Recently, several methods~\cite{liu2019knowledge,park2019relational,yu2019learning} tried to apply distillation to image embedding. They first trained a large teacher embedding model using existing methods and then distilled the knowledge learned by the teacher to a small student model. Instead of distilling the learned embedding vectors, the learned distances between embedding vectors are distilled.


\textbf{Unified Models} Gao \textit{et al.} trained a classification model with an extremely large number of classes~\cite{gao2017knowledge}. This was done by distilling the knowledge from a hierarchy of smaller models. Vongkulbhisal \textit{et al.}~\cite{vongkulbhisal2019unifying} developed a new task called Unifying Heterogeneous Classifiers (UHC) because of privacy considerations. In~\cite{vongkulbhisal2019unifying}, the scenario is that privately trained classification models can be shared but the private training data cannot be shared. They proposed a generalized distillation method to distill the knowledge in private models into a unified classification model. 
The task in this paper is similar to UHC in that both tasks aim to unify several models into one. This paper unifies image embedding models, while UHC unifies classification models.

\section{Method}
\label{sec:method}
We propose to distill the knowledge from multiple properly trained specialist MIR models to the universal multi-source MIR model. Let $D_i|_{i=1}^m$ denote the multiple medical image sources, where $m$ is the number of sources. For each source $D_i$, we first train a specialist MIR model $f_t^i$ using a state-of-the-art single-source method, \textit{i.e.} Multi-Similarity~\cite{wang2019multi}. Next, all the specialist MIR models are regarded as teachers to the universal multi-source MIR model. The distillation used in this paper is different from the previous methods~\cite{hinton2015distilling,park2019relational} because the multiple teachers have different functions. Therefore, we design a variation of the existing distillation methods. For the images from $D_i$, only $f_t^i$ is used to compute the distances between the images.

Relational knowledge distillation (RKD)~\cite{park2019relational,yu2019learning} is adopted for distillation from one teacher to the student. We present a brief review of RKD here. Let $\bm{x}_i$ and $\bm{x}_j$ denote two different images while $f_t(\cdot)$ and $f_s(\cdot)$ represent the teacher model and the student model, respectively. For simplicity, we define the embedding vectors for $\bm{x}$ computed by the teacher model and student model as $\bm{t} = f_t(\bm{x})$ and $\bm{s} = f_s(\bm{x})$, respectively. 
The relation distillation objective is to let the student mimic the learned distance between two embedding vectors from the teacher, which is 
\begin{equation}
    \mathcal{L}_{RKD} = l_\delta(d_t,d_s),
\end{equation}
where $l_\delta$ may be $\ell$1-distance loss or Huber loss~\cite{huber1992robust}. $d_t=\|\bm{t}_i-\bm{t}_j\|_2$ and $d_s=\|\bm{s}_i-\bm{s}_j\|_2$ are the pairwise distances between two images measured by the teacher model and student model, respectively. The Huber loss is defined as
\begin{equation}
    l_\delta(x,y)=\left\{ \begin{array}{ll}
    \frac{1}{2}(x-y)^2 & \text{for } |x-y|\leq 1,\\
    |x-y|-\frac{1}{2} & \text{otherwise}. 
    \end{array} \right.
\end{equation}

When the output dimensions of the teacher and student are largely different, the distance between two embeddings are normalized by dividing the mean distance in the batch in~\cite{park2019relational}. Mathematically, $d_t$ and $d_s$ are replaced by
\begin{equation}
d'_t=\frac{1}{\mu_t}\|\bm{t}_i-\bm{t}_j\|_2,\ 
d'_s=\frac{1}{\mu_s}\|\bm{s}_i-\bm{s}_j\|_2,
\label{eq:prop}
\end{equation}
where $\mu_t$ and $\mu_s$ are the mean distance between images in a batch.

A na\"ive way to sample the training images in a batch after fusing multiple sources is sampling without using the source information. The images from different sources are mixed together in a batch. The na\"ive sampling is not suitable for our multi-source MIR model training. Each specialist embedding model is trained with images in only one source, so the specialist can only be used to encode the images in that source. As a result, we are unable to compute the distances between images across sources. Based on this fact, we have each mini-batch only containing images from one source, which is named \textit{source-specific sampling}. The frequency of choosing one source to form a mini-batch is proportional to the number of images in that source. After determining which source to use, we follow the convention to choose the images inside a mini-batch. We randomly select $c$ classes and sample $k$ images for each class to form a mini-batch with size $c\times k$, where $k$ is set to $5$ in this paper. With the source-specific mini-batch sampling, we minimize the $\mathcal{L}_{RKD}$ between one specialist and the multi-source MIR model in each training iteration.


\section{Experiments}
\label{sec:exp}
We evaluate the proposed method by training a multi-source MIR model applicable to skin disease, x-ray, and retina images. Each one of Dermnet~\cite{liao2016skin}, Diabetic Retinopathy Detection Dataset\footnote{\url{https://www.kaggle.com/c/diabetic-retinopathy-detection/data}}, and ChestX-ray8~\cite{wang2017chestx} can be viewed as a source or domain. ImageNet~\cite{deng2009imagenet} is a large-scale nature image dataset and is used as an out-of-domain source for medical images. We first introduce the datasets used in this paper. Then we describe the experimental settings. In the end, the unification results are presented.

\begin{table*}[t]
\normalsize
  \caption{Performance comparisons of unifying Dermnet, Retina, and X-ray datasets. R1 refers to Recall@1, R2 refers to Recall@2, and R4 refers to Recall@4.}
  \label{tab:res}
  \centering
  \begin{tabular}{c|l|ccc|ccc|ccc|ccc}
    \toprule
     & & \multicolumn{3}{c|}{Dermnet} & \multicolumn{3}{c|}{Retina} & \multicolumn{3}{c|}{X-ray} & \multicolumn{3}{c}{Average} \\
     &  & R1 & R2 & R4 & R1 & R2 & R4 & R1 & R2 & R4 & R1 & R2 & R4 \\
    \midrule
    \parbox[t]{2mm}{\multirow{4}{*}{\rotatebox[origin=c]{90}{single sour.}}} 
      & Dermnet & \textbf{54.8} & \textbf{63.0} & \textbf{69.3} & - & - & - & - & - & - & -  & - & - \\
      & Retina & - & - & - & 47.0 & 64.9 & 78.7 & - & - & - & - & - & - \\
      & X-ray & - & - & - & - & - & -  & 26.3 & 37.9 & 52.5 & - & - & - \\
      & 3-way Concatenation & 50.1 & 58.5 & 65.8 & 46.0 & 64.4 & \textbf{80.1} & 24.8 & 38.1 & 53.7 & 40.3 & 53.7 & 66.5 \\
    \midrule
    \parbox[t]{2mm}{\multirow{4}{*}{\rotatebox[origin=c]{90}{multi-sour.}}} 
    & Multi-Similarity~\cite{wang2019multi} & 52.7 & 61.0 & 67.8 & 48.2 & 66.1 & 67.8 & 25.6 & 38.0 & 53.2 & 42.2 & 55.0 & 67.0 \\
    & Multi-Similarity+SS & 53.9 & 62.7 & \textbf{69.6} & 47.3 & 65.4 & 79.1 & 26.4 & 39.1 & 53.8 & 42.5 & 55.7 & 67.5 \\
    & Multi-Similarity+BAL & 53.1 & 61.6 & 68.5 & 47.4 & 65.2 & 79.0 & 25.7 & 38.9 & 54.4 & 42.1 & 55.3 & 67.3 \\
    & Ours & 54.0 & 61.7 & 67.2 & \textbf{50.1} & \textbf{67.4} & 80.0  & \textbf{28.7} & \textbf{41.0} & \textbf{55.5} & \textbf{44.3} & \textbf{56.7} & \textbf{67.6} \\
    \bottomrule
  \end{tabular}
\end{table*}

\begin{figure*}[t]
    \centering
\subfigure{
  \includegraphics[width=0.31\textwidth]{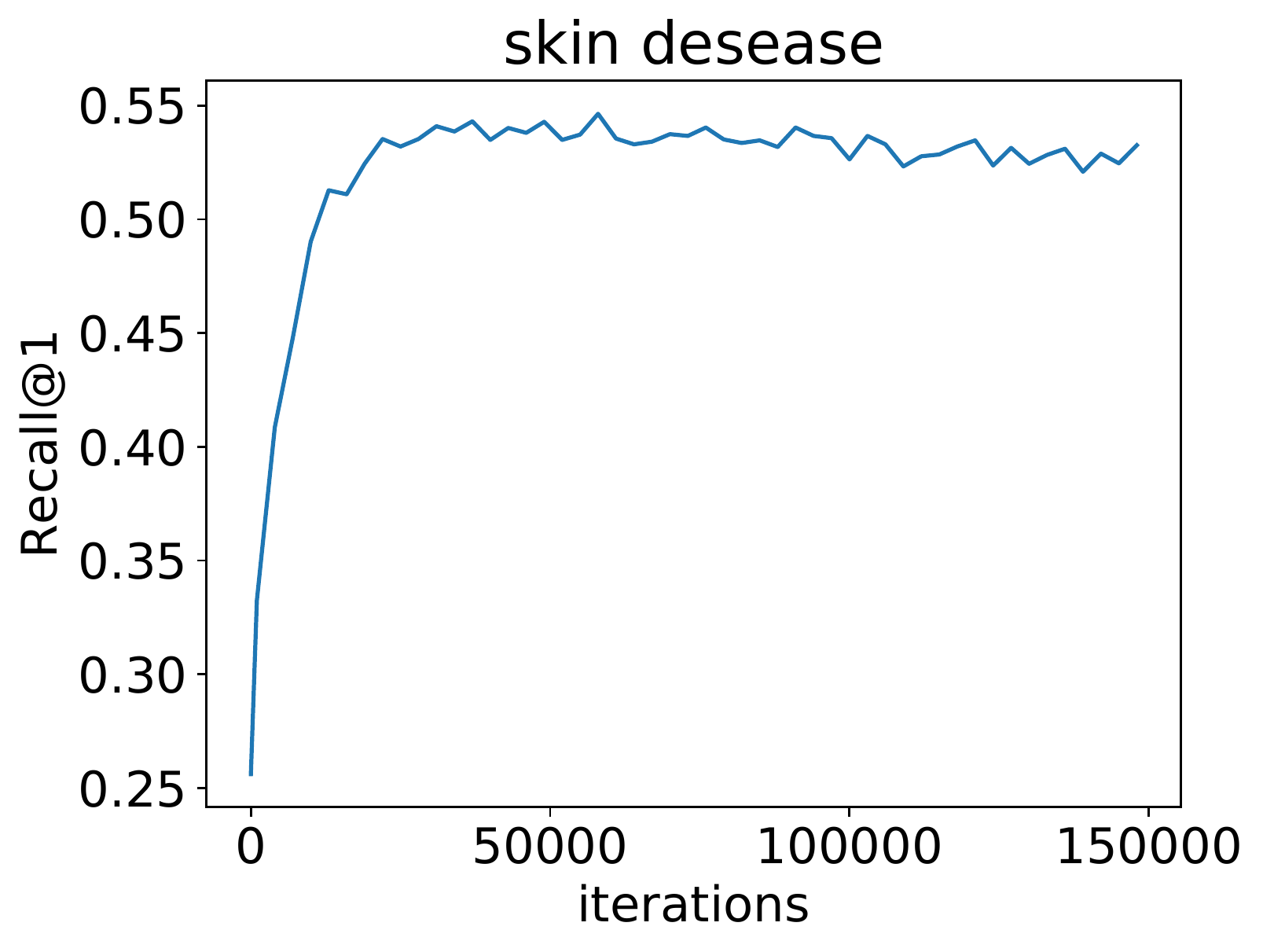}
}
\subfigure{
  \includegraphics[width=0.31\textwidth]{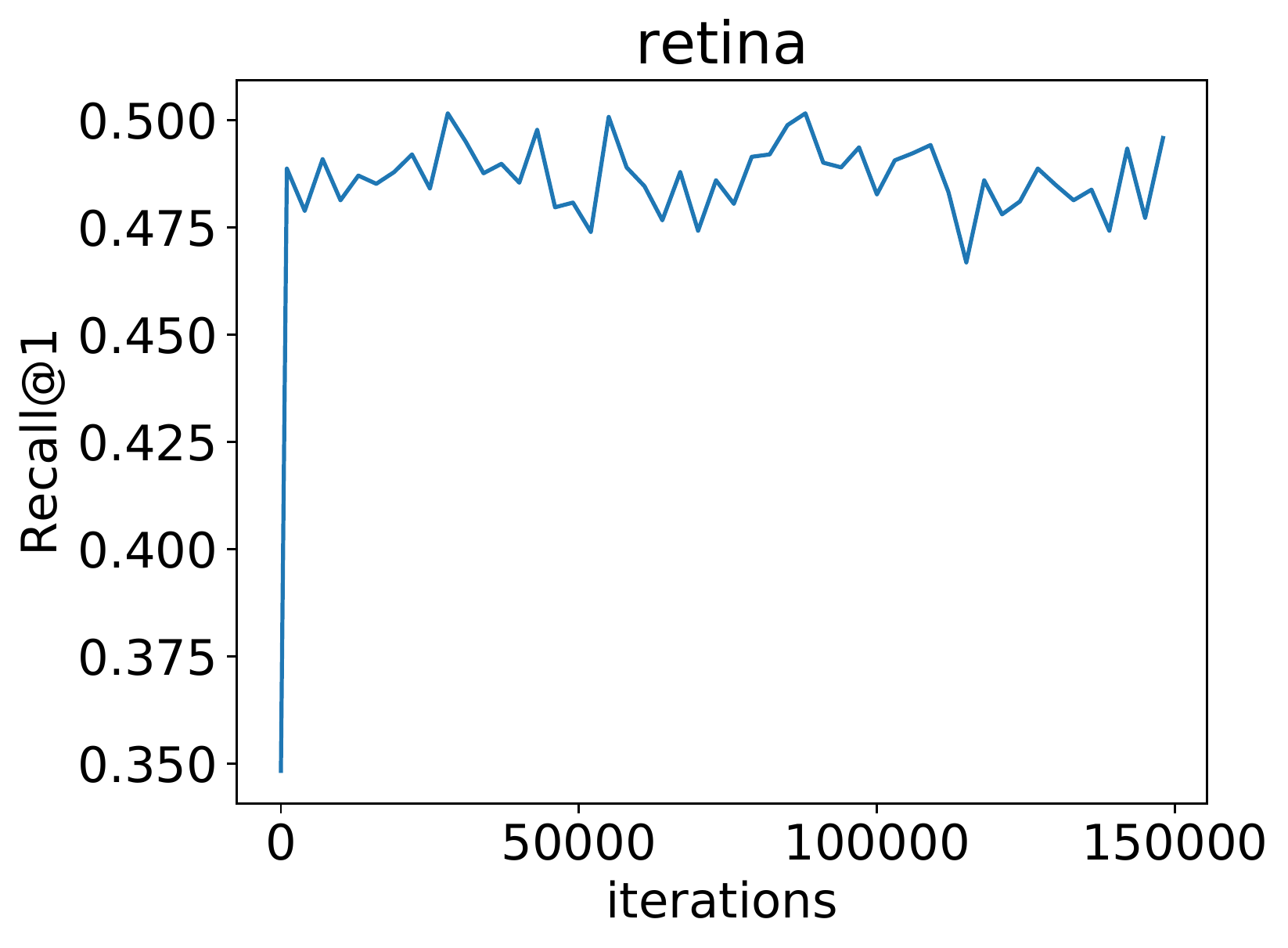}
}
\subfigure{
  \includegraphics[width=0.31\textwidth]{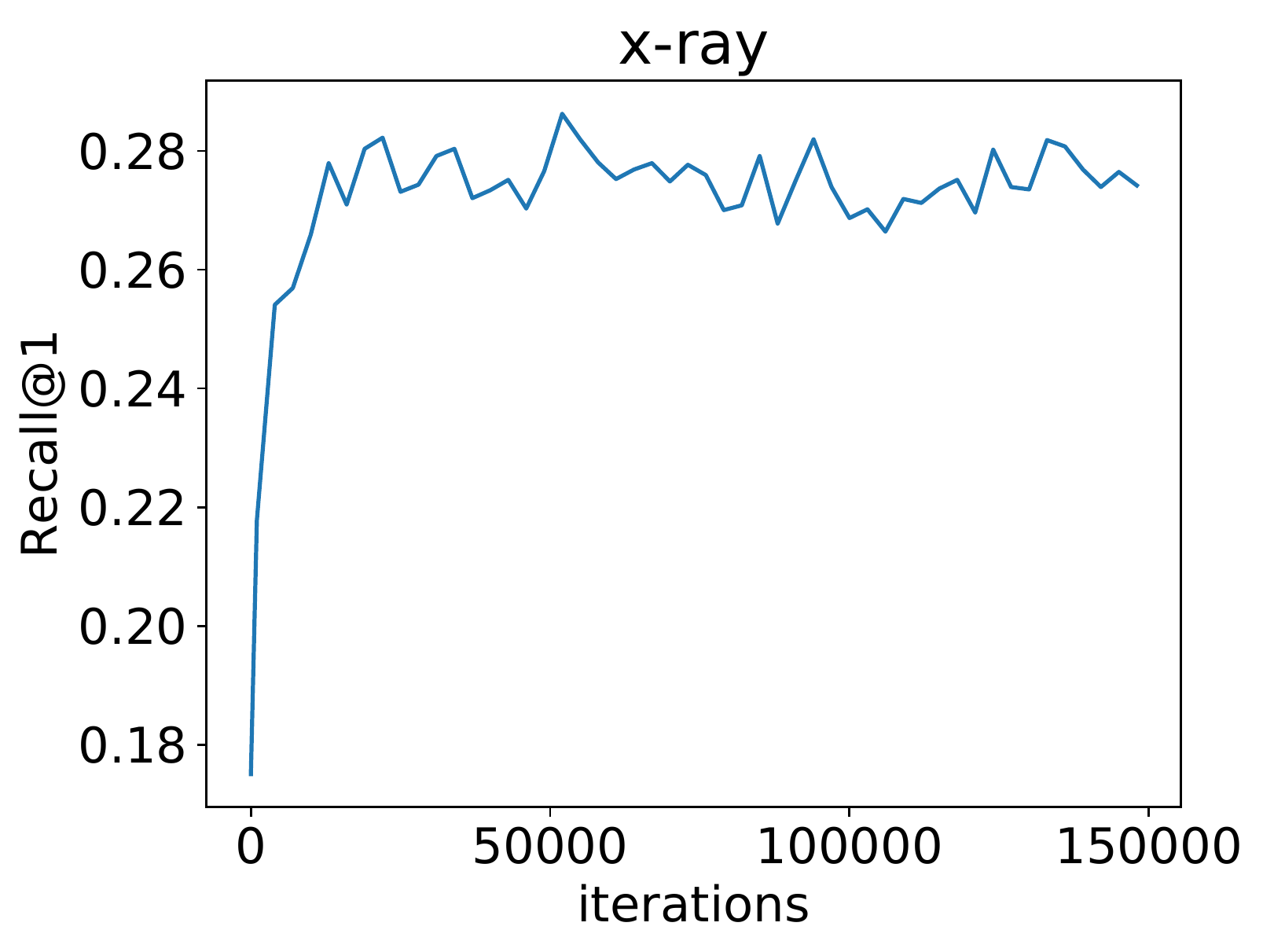}
}
    \caption{Visualization of the Recall@1 values when the skin disease, retina, and x-ray images are trained used the proposed distillation method.}
    \label{fig:recall_ours}
\end{figure*}

\subsection{Datasets}
\textbf{Dermnet} dataset contains 20,585 images in 335 categories. We split the images into training, validation, and testing split sets according to the ratio of 2:1:1. The resulted data split consists of 10,292 images for the training, 5,146 images for validation, and 5,147 images for testing.

\textbf{Diabetic Retinopathy Detection (Retina) Dataset} We only use the images released with labels. There are 35,126 images rated on a scale of 0 to 4. Because this dataset is highly imbalanced, we down-sample the most frequent labels so that the numbers of images belonging to the most frequent label and second most frequent label are the same. After downsampling, we have 14,608 images left. We then randomly split the dataset into training (7304 images), validation (3652 images), and testing (3652 images) sets. Some images in this dataset have very high resolution. To reduce the storage space consumption, we reduce the height and width by half if the width of an image is larger than 3,000 pixels.

\textbf{ChestX-ray8 (X-ray)} is originally a multi-class dataset. We first remove all the images having more than one labels to make a single-class dataset. After pruning, there are 91,324 images belonging to 15 classes. The data imbalance issue also exists for this dataset, so we down-sample the top two frequent classes, making the top three frequent classes have the same number of images. We split the remaining images into training, validation, and testing split sets according to the ratio of 2:1:1. In the end, there are 14923 images in the training split, 7461 images in the validation split, and 7462 images in the test split. All the images in this dataset are distributed in the PNG format, which takes large storage space. We convert the images into JPEG format using Python Imaging Library (PIL)\footnote{\url{https://pillow.readthedocs.io/en/stable}}, keeping a image quality (95).

\textbf{ImageNet} refers to ILSVRC2012~\cite{deng2009imagenet}. It is widely used for object recognition. In this paper, we use ImageNet as an out-of-domain source for MIR. The $1,281,167$ training images are used for training and the $50,000$ validation images are used for testing.

\begin{figure*}[t]
    \centering
\subfigure{
  \includegraphics[width=0.31\textwidth]{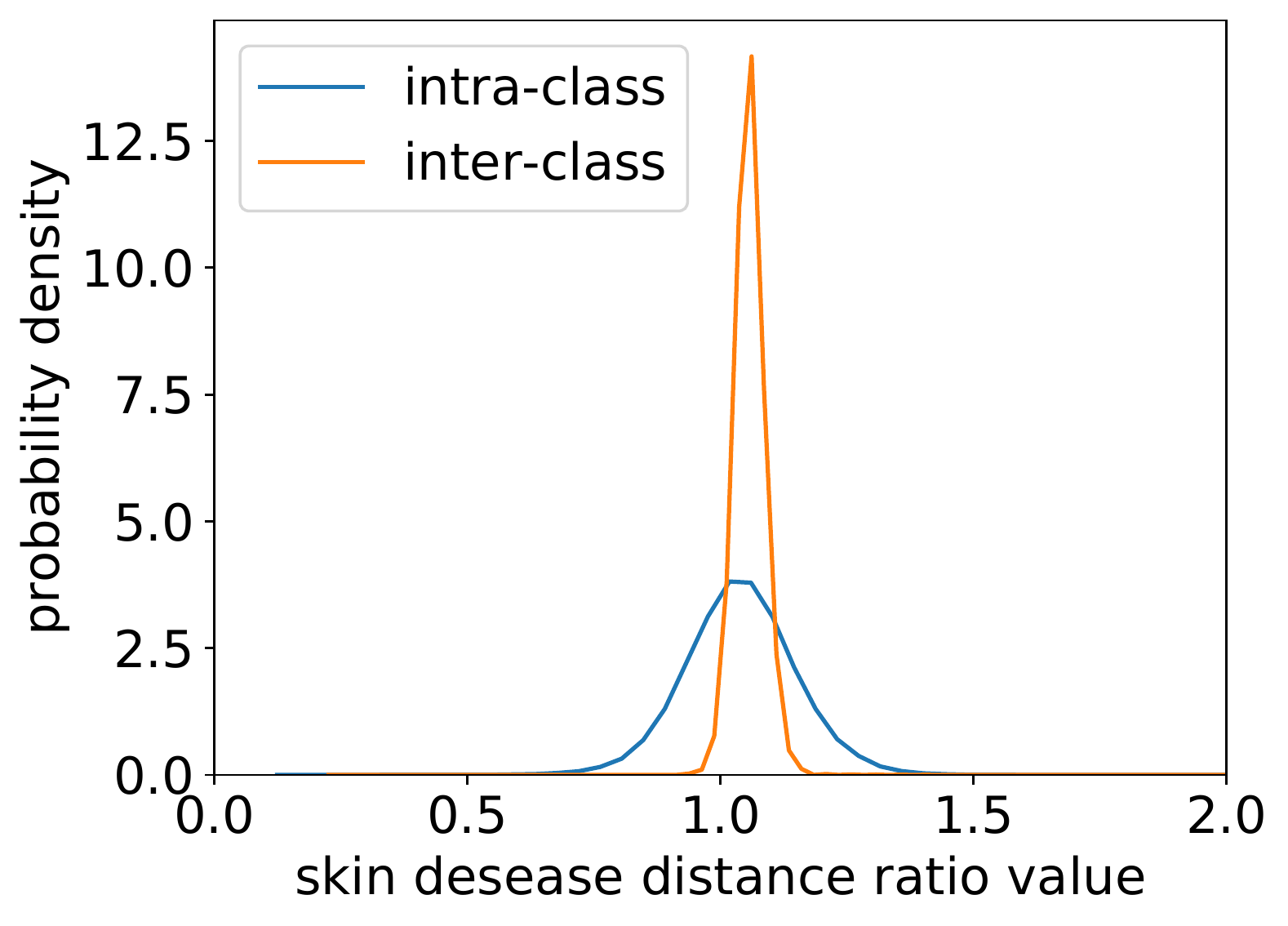}
}
\subfigure{
  \includegraphics[width=0.31\textwidth]{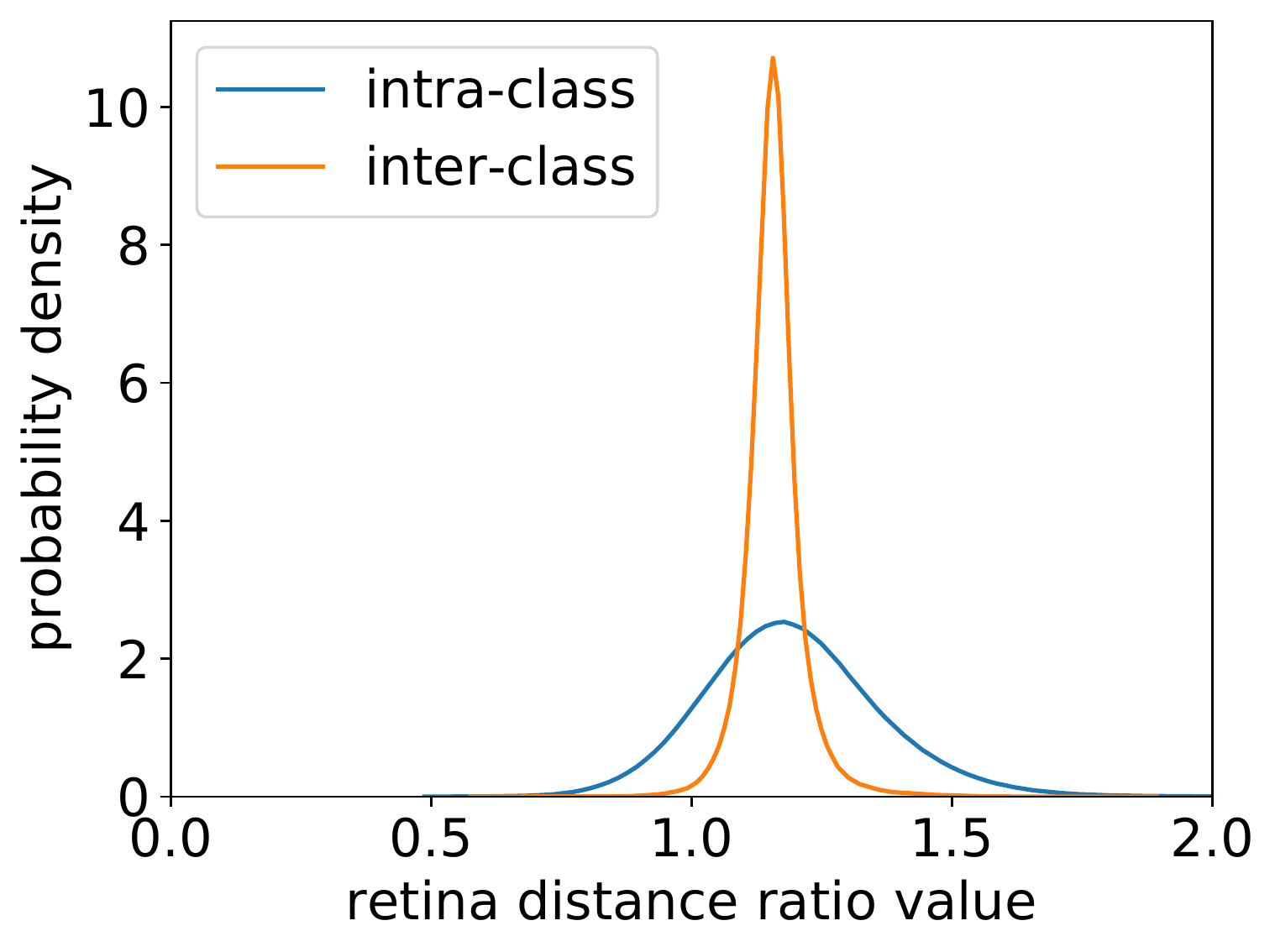}
}
\subfigure{
  \includegraphics[width=0.31\textwidth]{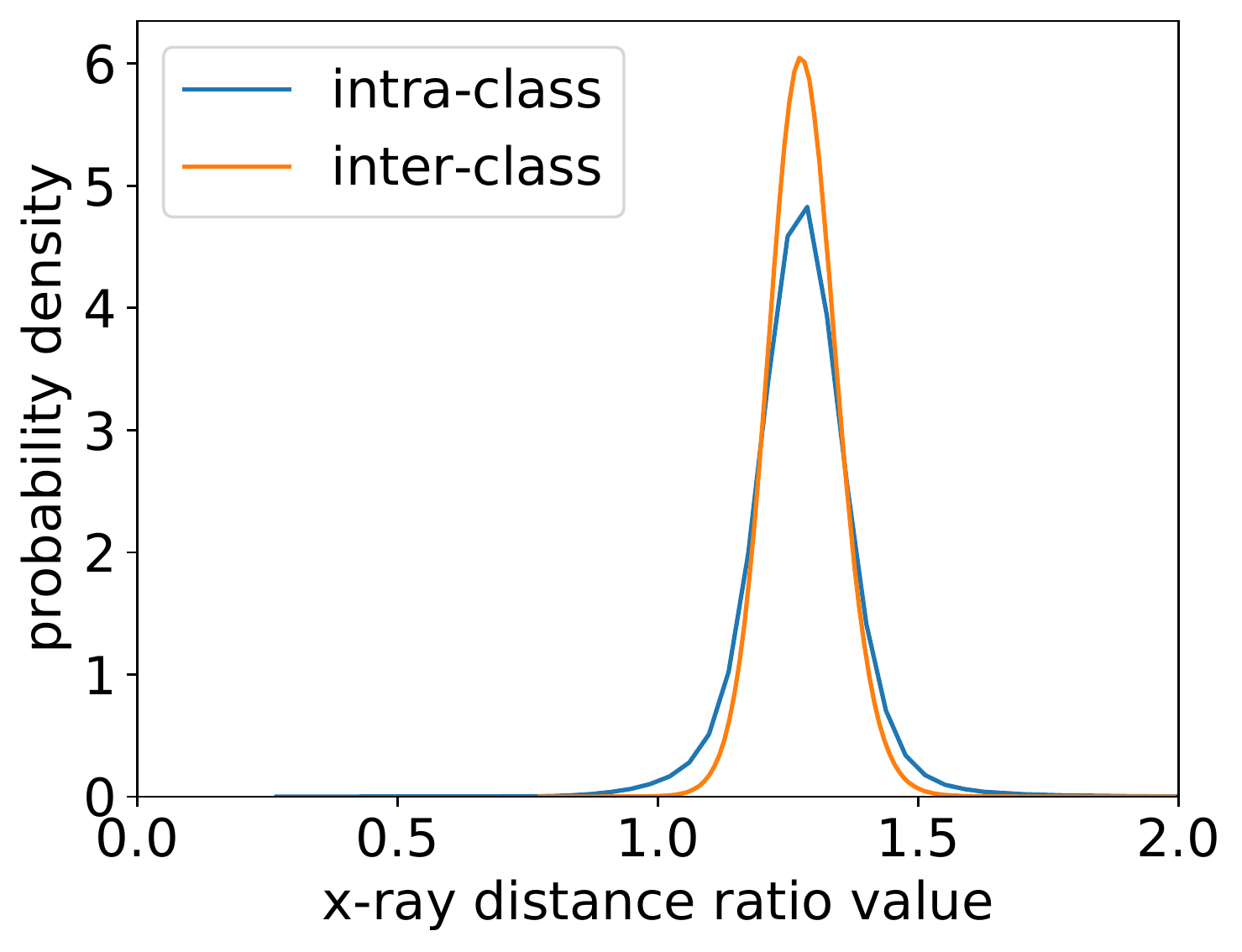}
}
    \caption{Visualization of the distribution of the ratio of distances measured by the multi-source MIR model and the specialist MIR models.}
    \label{fig:ratio}
\end{figure*}

\subsection{Settings}
In all the experiments, we use ResNet50~\cite{he2016deep} pre-trained for ImageNet classification task as the backbone CNN. 
The training images are resized to $256\times 256$ and then randomly cropped to $224\times 224$. Random horizontal flipping is also used for data augmentation. Central crop is used for the test images. We add a fully-connected layer to project the 2048-dim ResNet50 output into a 128-dim embedding vector and further normalize the embedding vector to unit-length. To train a specialist, we use Multi-Similarity~\cite{wang2019multi}. The batch size is set to $130$ and the Adam optimizer with the learning rate $1e^{-5}$ is used for training. The Recall@k are reported for performance comparison. The proposed method is implemented in  Tensorflow~\cite{tensorflow2015-whitepaper}. Nvidia GTX 1080Ti GPUs are used to train the embedding models.

We design four baseline methods for comparison. In the ``Concatenation'' baseline, we concatenate the embedding vectors extracted by the trained specialists, which can be considered as an ensemble method. To make the comparison fair, we use PCA to project the concatenated embedding into 128-dim. The remaining three baselines can be summarized by fusing the training data and then training with single source methods. The difference between these three baselines lies in the sampling method: 1) na\"ive sampling; 2) source specific (SS) sampling introduced in Sec.~\ref{sec:method}; and 3) source-balanced (BAL) sampling. In na\"ive sampling, the training images from different datasets are mixed in each training batch. BAL sampling is based SS sampling, but the datasets are sampled with an equal probability regardless of the number of training images from each dataset.

\begin{figure}[t]
    \centering
\subfigure{
  \includegraphics[width=\columnwidth]{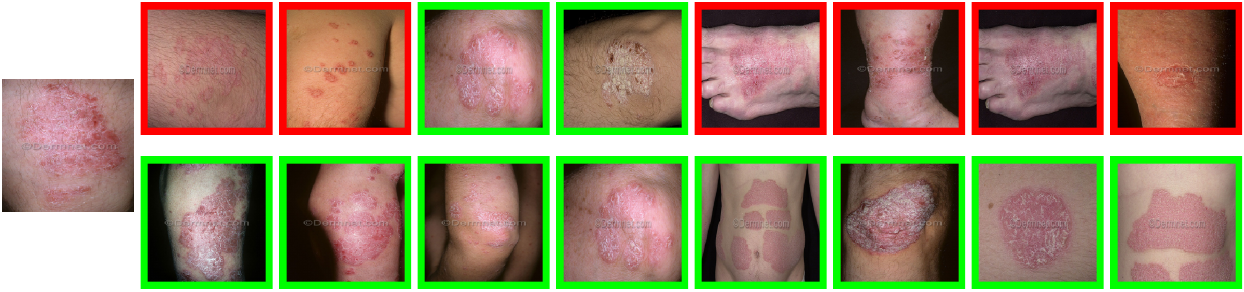}
}
\subfigure{
  \includegraphics[width=\columnwidth]{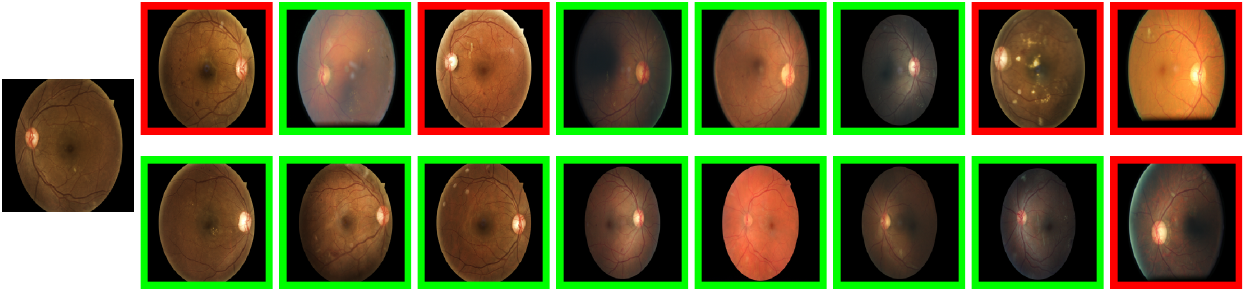}
}
\subfigure{
  \includegraphics[width=\columnwidth]{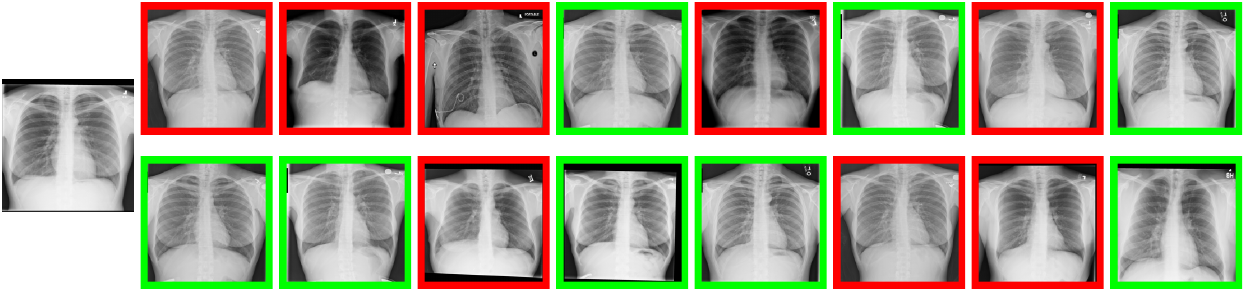}
}
    \caption{Qualitative results of MIR. In the top, middle and bottom sub-figures, we show the qualitative retrieval results of Derment, Retina and X-ray datasets, respectively. The query image is shown on the left of each sub-figure. On the right of each sub-figure, there are two rows. The first row shows the top-8 retrieval results by the specialist MIR model and the second row shows the top-8 retrieval results by the multi-source MIR model. The correct retrieval images are marked by green and the incorrect images are marked by red.}
    \label{fig:vis}
\end{figure}

\begin{table*}[t]
\normalsize
  \caption{Performance comparisons of unifying Dermnet, Retina X-ray, and ImageNet datasets. R1 refers to Recall@1, R2 refers to Recall@2, and R4 refers to Recall@4.}
  \label{tab:res-in}
  \centering
  \begin{tabular}{c|l|ccc|ccc|ccc|ccc}
    \toprule
     & & \multicolumn{3}{c|}{Dermnet} & \multicolumn{3}{c|}{Retina} & \multicolumn{3}{c|}{X-ray} & \multicolumn{3}{c}{Average} \\
     &  & R1 & R2 & R4 & R1 & R2 & R4 & R1 & R2 & R4 & R1 & R2 & R4 \\
    \midrule
    \parbox[t]{2mm}{\multirow{2}{*}{\rotatebox[origin=c]{90}{sing.}}} 
    & ImageNet & 22.1 & 29.0 & 36.5 & 35.1 & 56.2 & 76.3 & 15.4 & 26.7 & 43.0 & 24.2 & 37.3 & 51.9 \\
    & 4-way Concatenation & 45.0 & 53.9 & 62.6 & 45.9 & 65.0 & 80.1 & 23.0 & 35.5 & 52.1 & 38.0 & 51.5 & 64.9 \\
    \midrule
    \parbox[t]{2mm}{\multirow{4}{*}{\rotatebox[origin=c]{90}{multi-sour.}}} 
    & Multi-Similarity~\cite{wang2019multi} & 51.9 & 60.7 & 67.0 & 48.5 & 66.5 & 80.0 & 24.9 & 37.4 & 52.5 & 41.8 & 54.9 & 66.5 \\
    & Multi-Similarity+SS & 53.1 & 61.3 & 67.7 & 47.7 & 65.6 & \textbf{80.2} & 24.9 & 38.4 & 53.3 & 41.9 & 55.1 & 67.0 \\
    & Multi-Similarity+BAL & 51.3 & 59.5 & 64.9 & 46.7 & 64.0 & 80.0 & 25.6 & 37.7 & 52.9 & 41.2 & 53.7 & 65.8 \\
    & Ours & \textbf{54.7} & \textbf{62.9} & \textbf{68.9}  & \textbf{50.7} & \textbf{67.9} & 79.5 & \textbf{27.9} & \textbf{40.4} & \textbf{55.2} & \textbf{44.4} & \textbf{57.1} & \textbf{67.9} \\
    \bottomrule
  \end{tabular}
\end{table*}

\subsection{Multi-source MIR Results}
The results of all methods are listed in Table~\ref{tab:res}. The recall values are computed on the testing splits of the three datasets. By comparing the results in the Retina column, we can find that all the baselines and the proposed method have better performance than the Retina specialist, which may be because the Retina source benefits from the other two sources. This finding is consistent with \cite{zhou2019models}. The ``SS'' sampling method solves the ineffective triplet problem and achieves better averaged recall than na\"ive sampling. The ``Concatenation'' baseline achieves the worst performance, which may be because two of the three specialist models are out-of-domain. The proposed method in this paper achieves better performance than all the baselines, confirming the advantage of universal embedding for multi-source MIR with a single model.

In Fig.~\ref{fig:recall_ours}, we show the Recall@1 of the multi-source MIR model. We can find that the early over-fitting problem no longer happens because of training by distillation. Given pairs of images, the distribution of the ratios of the distances measured by the multi-source MIR model and the specialist MIR models is shown in Fig.~\ref{fig:ratio}. For all the three sources, we can find that the mean values of the ratio are greater than 1, meaning that the distances become larger when measured by the multi-source MIR model. For Dermnet and Retina sources, the variances of the inter-class distance ratios are smaller than the intra-class distance ratios. Some of the qualitative image retrieval results are shown in Fig.~\ref{fig:vis}. In the top, middle and bottom sub-figures of  Fig.~\ref{fig:vis}, we show the retrieval results of Derment, Retina and X-ray datasets, respectively. On the right of each sub-figure, the first row shows the top-8 retrieval results by the specialist MIR model and the second row shows the top-8 retrieval results by the multi-source MIR model. The correct retrieval images are marked by green boxes and the incorrect images are marked by red boxes.

\subsection{MIR with ImageNet as an Out-of-domain Source}
In this experiment, we use nature images from the ImageNet dataset as an out-of-domain source for MIR. ImageNet contains $1,000$ categories over a broad range. The total number of training images in ImageNet is 91, 175, 86 times larger than DermNet, Retina and X-ray dataset, respectively. If we simply sample the mini-batch proportional to the number of training images, it is very unlikely to select the images from the medical dataset sources. So we increase the probability of choosing the training images from medical sources by a hundred times. The result sampling policy is used for all of ``Multi-Similarity'', ``Multi-Similarity+SS'', and ``Ours'' method in this section. Since no early overfitting happens to the ImageNet dataset, it is not necessary to use distillation for ImageNet images. We directly use the Multi-Similarity loss instead of the distillation loss on the ImageNet images during training.

The results are reported in Table~\ref{tab:res-in}. In the first row, we apply the ImageNet specialist model on all three medical sources. It is not surprising  that the ImageNet model achieves low performance on medical images. ``Ours'' method achieves the best performance, which confirms the effectiveness of distillation. The average R1 of ``Ours'' in Table~\ref{tab:res-in} is only slightly higher than that value in Table~\ref{tab:res}, which may be because the pre-trained ResNet already contains the knowledge from the ImageNet dataset.

\section{Conclusions}
In this paper, we investigate the novel and useful problem of training a universal multi-source MIR model to have good performance on multiple sources. We first show that fusing the training data from all the sources and training with single-source methods cannot solve this problem. We then propose to distill the knowledge learned by properly trained specialist models into a universal multi-source MIR model. The experimental results on several diverse medical image sources have shown the effectiveness of the proposed universal model.






\bibliographystyle{IEEEtran}
\bibliography{IEEEexample}
%

\end{document}